\documentclass[letterpaper, 10 pt, conference]{ieeeconf}  

\IEEEoverridecommandlockouts                              

\usepackage{graphics} 
\usepackage{epsfig} 
\usepackage{mathptmx} 
\usepackage{times} 
\usepackage{amsmath} 
\usepackage{amssymb}  
\usepackage{amsfonts}

\usepackage{subfig}
\usepackage{hyperref}
\usepackage{dblfloatfix}

\title{\LARGE \bf
Tackling Real-World Autonomous Driving using Deep Reinforcement Learning
}

\author{Paolo Maramotti$^{1}$, Alessandro Paolo Capasso$^{2}$, Giulio Bacchiani$^{2}$ and Alberto Broggi$^{2}$
\thanks{$^{1}$Paolo Maramotti is with Vislab - University of Parma, Parma, Italy
        {\tt\small paolo.maramotti@unipr.it}}%
\thanks{$^{2}$Alessandro Paolo Capasso, Giulio Bacchiani, Alberto Broggi are with Vislab srl, an Ambaraella Inc. company - Parma, Italy
        {\tt\small acapasso@ambarella.com, gbacchiani@ambarella.com, broggi@vislab.it}}%
}

\begin{document}

\maketitle
\thispagestyle{empty}
\pagestyle{empty}

\begin{abstract}

In the typical autonomous driving stack, planning and control systems represent two of the most crucial components in which data retrieved by sensors and processed by perception algorithms are used to implement a safe and comfortable self-driving behavior. In particular, the planning module predicts the path the autonomous car should follow taking the correct high-level maneuver, while control systems perform a sequence of low-level actions, controlling steering angle, throttle and brake. In this work, we propose a model-free Deep Reinforcement Learning Planner training a neural network that predicts both acceleration and steering angle, thus obtaining a single module able to drive the vehicle using the data processed by localization and perception algorithms on board of the self-driving car. In particular, the system that was fully trained in simulation is able to drive smoothly and safely in obstacle-free environments both in simulation and in a real-world urban area of the city of Parma, proving that the system features good generalization capabilities also driving in those parts outside the training scenarios.
Moreover, in order to deploy the system on board of the real self-driving car and to reduce the gap between simulated and real-world performances, we also develop a module represented by a tiny neural network able to reproduce the real vehicle dynamic behavior during the training in simulation. 

\end{abstract}

\section{INTRODUCTION}

In recent years, the technological progress in mechanical, electronic and IT areas has led industrial and academic sectors to invest resources in the autonomous driving field. 
In the last decades, excellent progress has been conducted for increasing the level of vehicle automation, starting from simple, rule-based approaches \cite{marsden2001towards} to implementing smart systems based on Artificial Intelligence (\cite{chen2015deepdriving}, \cite{ma2020artificial}).
In particular, these systems aim at solving the main limitations of the rule-based approaches that are the lack of negotiation and interaction with other road users and the poor understanding of the dynamics of the scenario.

\begin{figure}
    \centering
    \includegraphics[width=0.48\textwidth]{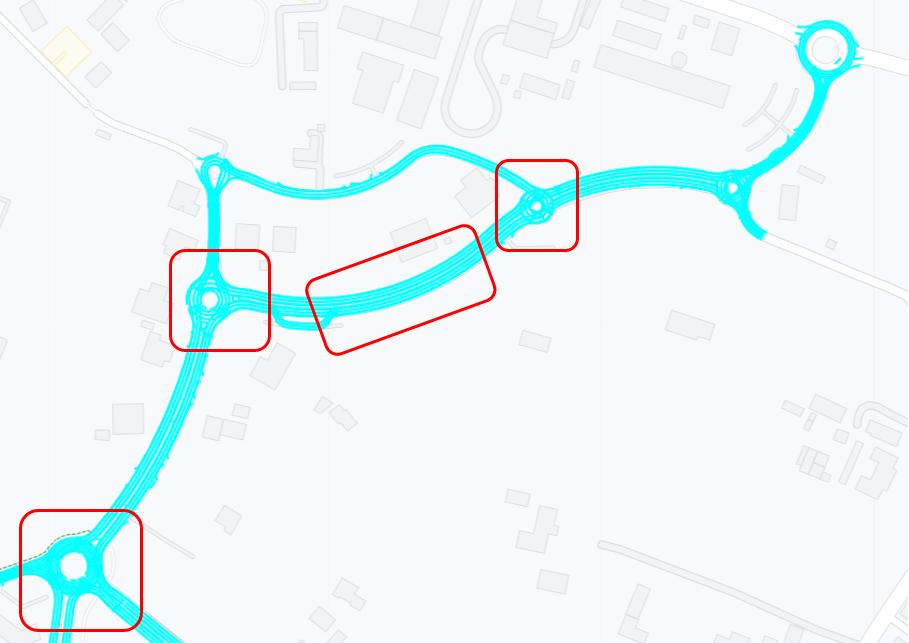}
    \caption{Mapped area of a neighborhood of Parma. The four red rectangles contain the scenarios used for training agents. The whole light blue area was used for testing.}
    \label{fig:hd_map}
\end{figure}

In this paper, we focus on the development of a Deep Reinforcement Learning (DRL) \cite{arulkumaran2017deep} planner, able to drive the vehicle safely and comfortably in obstacle-free environments, using a neural network that predicts both the acceleration and the steering angle.\\
Indeed, Reinforcement Learning (RL) \cite{sutton2018reinforcement} is widely used to solve tasks using both discrete control space output, such as Go \cite{silver2017mastering}, Atari games \cite{mnih2015human} or chess \cite{silver2018general}, and continuous control space \cite{duan2016benchmarking}, as autonomous driving \cite{kiran2021deep}.\\
In particular, RL algorithms are widely used in the autonomous driving field for the development of decision-making and maneuver execution systems like lane change (\cite{wang2018reinforcement}, \cite{hoel2018automated}, \cite{mirchevska2018high}), lane keeping (\cite{sallab2016end}, \cite{feher2018q}), overtaking maneuvers \cite{kaushik2018overtaking}, intersection and roundabout handling (\cite{garcia2019autonomous}, \cite{capasso2020intelligent}) and many others.\\
Starting from the delayed version of Asynchronous Advantage Actor Critic (A3C) algorithm (\cite{mnih2016asynchronous}, \cite{capasso2020simulation}, \cite{capasso2021end}), we implemented a Reinforcement Learning planner training agents in a simulator based on High Definition Maps (HD Maps \cite{liu2020high}) developed internally by the research team. In particular, we trained the model predicting continuous actions related to the acceleration and steering angle, and testing it on board of a real self-driving car on an entire urban area of the city of Parma (Fig. \ref{fig:hd_map}). 

Moreover, we demonstrate that the system achieved good generalization performances both in simulation and in real world tests, showing that the module is able to drive smoothly and safely also in those areas not included in the training scenarios.
Furthermore, in order to obtain agents that behave as similar as possible to the real self-driving car, a simple model based on a neural network (called \textit{deep$\_$response}) has been developed to reproduce in simulation the real self-driving vehicle dynamic behavior. In this way, we reduce the gap between the simulated agent and the real vehicle behavior, proving that the \textit{deep$\_$response} module is essential to obtain a comfort and safe driving style in real-world scenarios.

Finally, since RL algorithms take millions of episodes before converging to an optimal solution, especially in complex tasks as in the autonomous driving field, we decided to perform a pre-training using Imitation Learning (IL) \cite{hussein2017imitation} that allow us to overcome this limit. 
In particular, we collected a dataset using agents that drive in the scenarios following deterministic rules. \\
Therefore, we exploited the ability of Imitation Learning to converge more quickly to an optimal solution to obtain network parameters as initial weights for the RL training. In this way we were able to drastically reduce the time needed to obtain an optimal policy.

\section{RELATED WORKS}
The development of the autonomous driving systems combined with the progress of neural networks, has increasingly turned towards learn-based systems.
\\
In \cite{sallab2017deep} the authors proposed a framework for an end-to-end Deep Reinforcement Learning \cite{arulkumaran2017deep} pipeline for autonomous driving which integrates RNNs (Recurrent Neural Network) \cite{rumelhart1985learning} to account for Partially Observable MDP (Markov Decision Process); they also integrate an attention models into the framework scenarios and test it in TORCS simulator \cite{wymann2000torcs}. Since 2015, when DeepMind published its work on the Deep Q-Network \cite{mnih2015human} in which, through Reinforcement Learning, they solve many Atari games, some researchers have begun to test such approach in other areas as well. In the autonomous driving field, in \cite{wolf2017learning} the authors presented a RL approach that uses Deep Q-Networks to drive a vehicle in a 3D physics simulator in an end-to-end manner. Similarly, based on Deep Q-Networks and merging it with Deterministic Policy Gradient (DPG), in \cite{lillicrap2015continuous} was proposed Deep Deterministic Policy Gradient (DDPG). This is a method which can be used to solve the control problem in the high-dimensional and continuous action space and was also tested in the autonomous driving field in TORCS simulator \cite{wymann2000torcs}.

The first work that showed that DRL is a viable approach to autonomous driving in real world is \cite{kendall2019learning}. They developed a system able to perform the lane following maneuver using a single monocular camera. The system differs from the others in the literature because both training and testing are done directly on the vehicle and not only in simulation.

In this paper we used a delayed version of the original A3C \cite{mnih2016asynchronous} called Delayed-A3C (D-A3C).
This algorithm was previously developed and used in \cite{capasso2020simulation} and \cite{capasso2021end} where it has been shown that it allows to achieve better results than A3C.
In D-A3C configuration, each agent begins the episode with a local copy of the latest version of the global network, while the system collects all the contribution of the actors; the agent updates their local copy of the network at fixed time intervals but all the updates are sent to the global network only at the end of the episode, while in the classical A3C algorithm this exchange is performed at fixed time intervals.

Another approach widely used for the development of planning and control systems is Imitation Learning, which is a supervised learning technique that allows to learn a policy by demonstration, or rather by mimicking the behavior collected in the dataset. For this purpose, NVIDIA (\cite{bojarski2016end}, \cite{bojarski2017explaining}) developed an end-to-end driving system using deep convolutional neural networks. They trained the system to map images captured from a single front camera directly into steering commands. 
This was followed by many other works such as \cite{chen2019deep} in which they do not use as input for the system the raw data taken directly from the sensors, but process them to obtain bird's-eye-views that contain a more concise and generalized representation of the data. In this way, unnecessary information can be discarded and only useful information for decision making and planning are kept.\\
Equally interesting are the works \cite{dosovitskiy2017carla} and \cite{codevilla2018end}, where CARLA simulator was created and on which they tested their conditional imitation learning algorithms, developed to train an agent to drive following the road and avoiding obstacles, based on the behavior of a human. \\
The main advantage of IL respect to RL is that is not necessary to manually design the policy model or cost function.
However, IL has the main limitation of requiring a large amount of data in order to avoid overfitting and at the same time it is difficult to create datasets containing all the possible situations a vehicle could encounter. 
\begin{figure*}
    \centering
    \subfloat[Top-view of the training scenario]{\includegraphics[width=0.28\textwidth]{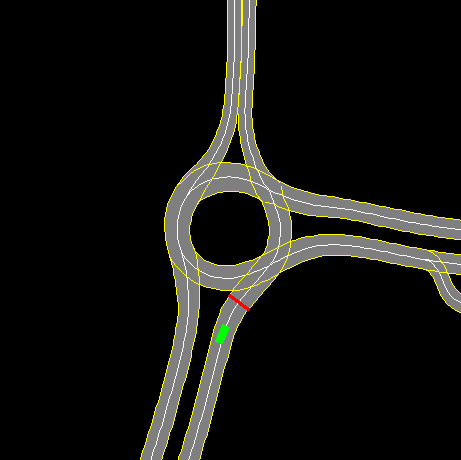}\label{scen_a}}
    
    \subfloat[Surrounding view]{\includegraphics[width=0.14\textwidth]{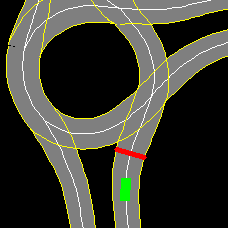}\label{scen_b}}\hspace{1.0mm}
    \subfloat[Obstacles]{\includegraphics[width=0.14\textwidth]{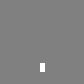}\label{scen_c}}\hspace{1.0mm}
    \subfloat[Navigable space]{\includegraphics[width=0.14\textwidth]{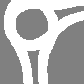}\label{scen_d}}\hspace{1.0mm}
    \subfloat[Path]{\includegraphics[width=0.14\textwidth]{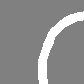}\label{scen_e}}\hspace{1.0mm}
    \subfloat[Stop line]{\includegraphics[width=0.14\textwidth]{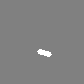}\label{scen_f}}
    \caption{Synthetic representation of training scenario (Fig. \ref{scen_a}) included in the mapped area of Fig. \ref{fig:hd_map}. Fig. \ref{scen_b} represents the $50\times50$ meters surrounding perceived by the agent, that is split in four channels: obstacles (Fig. \ref{scen_c}), navigable space (Fig. \ref{scen_d}), path (Fig. \ref{scen_e}) and stop line (Fig. \ref{scen_f}). }
    \label{fig:scenario}
\end{figure*}

\section{NOTATION}
A typical Reinforcement Learning algorithm involves the interaction between at least one agent and its environment. Through the trials and errors mechanism, it learns to behave in order to achieve its goal by receiving a reward function to evaluate the goodness of the chosen action. In general, at time $t$ the agent performs an action $a_t$ observing a state $s_t$; as a consequence of such action it will receive a reward signal $r_t$ finding itself in a new state $s_{(t+1)}$.\\
The RL problem can be defined as a Markov Decision Process (MDP), $M = (S, A, P, r, \gamma)$, where $S$ is a set of states, $A$ is a set of discrete or continuous actions, $P$ is the state transition probability $P(s_{t+1} | s_t, a_t)$, $r$ represents the reward function and $\gamma$ is the discount factor ($[0,1]$) that modulates the importance of future rewards.
Therefore, the goal of a Reinforcement Learning agent is to maximize the expected return $R_t = \begin{matrix} \sum_{t}^T r_t + \gamma r_{t+1} + \cdots + \gamma^{T-t} r_T \end{matrix}$ (where $T$ represent the time instant of the terminal state).

In this paper we use the delayed version of Asynchronous Advantage Actor Critic (D-A3C) that belongs to the family of the so called Actor-Critic algorithms \cite{mnih2016asynchronous}. In particular, it is composed by two different entities: the \textit{actor} and the \textit{critic}. The purpose of the \textit{actor} is to choose the action that the agent has to perform, while the \textit{critic} estimates the state-value function, namely the goodness of a specific state for the agent. In other words, the \textit{actor} is a probability distribution over actions $ \pi(a|s;\theta^\pi)$ (where $\theta$ are the network parameters) and the \textit{critic} is estimates the state-value function $v(s_t;\theta^v) = \mathbb{E} (R_t|s_t)$.

\section{TRAINING SETTINGS}

\subsection{Environment}
For the development of such work, a synthetic simulator has been implemented using High Definition Maps both developed internally by the team; an example of a scenario is illustrated in Fig. \ref{scen_a}. This represents a portion of the mapped area (Fig. \ref{fig:hd_map}) in which we tested our system on board of the real self-driving car.
The Fig. \ref{scen_a} shows a full view of such training scenario while the Fig. \ref{scen_b} the surrounding view perceived by the agent, corresponding to an area of $50\times50$ meters. This is divided into four channels: the obstacles (Fig. \ref{scen_c}) that at this stage of the work are only represented by the ego vehicle, the navigable space (Fig. \ref{scen_d}), the path (Fig. \ref{scen_e}) that the agent should follow and that is randomly calculated at the beginning of the episode and, finally, the stop line (Fig. \ref{scen_f}).
The use of HD Maps in the simulator allows us to retrieve several information about the external environment like the positions or the number of lanes, the road speed limits and much more. \\
In this phase of the work, we focused on the achievement of a smooth and safely driving style, and for this reason we trained our agents in static scenarios, without including obstacles or other road users, learning to following the route and respecting the speed limits.

\subsection{Neural Network}
\label{nn}

\begin{figure*}
    \centering
    \includegraphics[width=0.8\textwidth]{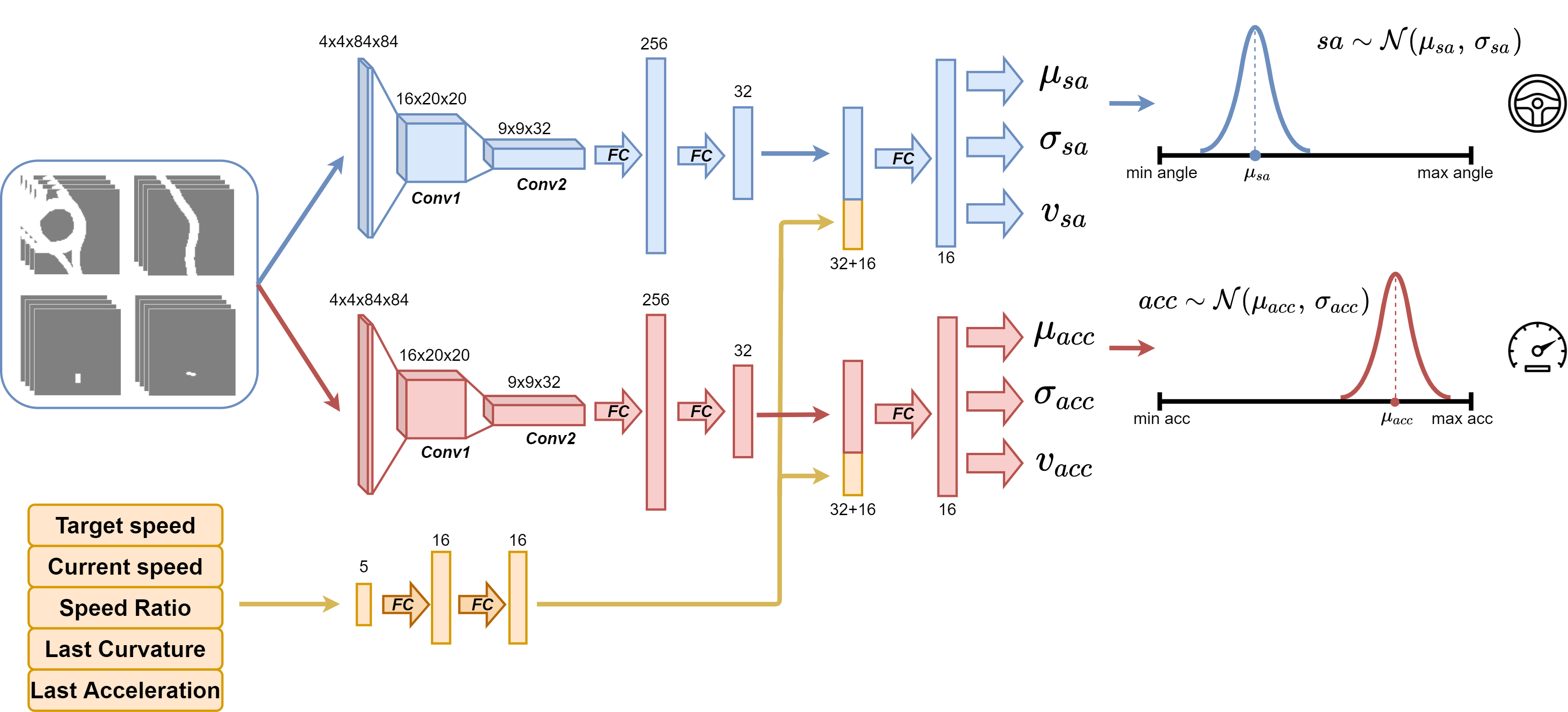}
    \caption{The neural network architecture used for training agents to drive
following their paths and observing the road speed limits.  The net is composed by two sub-modules that predict acceleration and steering angle, sampled from two Gaussian distribution ($acc = \mathcal{N}(\mu_{acc}, \sigma_{acc})$, $sa = \mathcal{N}(\mu_{sa}, \sigma_{sa})$). 
Both the sub-modules receive the same five scalar parameters (Target speed, Current speed, Speed ratio, Last curvature, Last acceleration) and the same four visual input, each one composed by the last four sequential frame images  $84\times84$ pixels in order to give the agent a history of the past states and, consequently understand the dynamic of the environment. }
    \label{fig:network}
\end{figure*}

Agents are trained using the neural network illustrated in Fig. \ref{fig:network}, that predicts steering angle and acceleration set points every 100 ms. It is divided into two main sub-modules: the first one able to define the steering angle $sa$ and the second one for the acceleration $acc$. 
The input of both sub-modules is represented by 4 channels (navigable space, path, obstacles and stop line) corresponding to the surrounding view of the agent (Fig. \ref{fig:scenario}). Each visual input channel contains 4 images of $84\times84$ pixels in order to give the agent an history of the past states. Together with this visual input, the network receives 5 scalar parameters including the target speed (the road speed limit), the current speed of the agent, the ratio between the current speed and the target speed, and the last actions related to the steering angle and the acceleration. \\
In order to guarantee exploration, the output of the two sub-modules is sampled from two Gaussian distributions to obtain relatively acceleration ($acc = \mathcal{N}(\mu_{acc}, \sigma_{acc})$) and steering angle ($sa = \mathcal{N}(\mu_{sa}, \sigma_{sa})$). 
The standard deviations $\sigma_{acc}$ and $\sigma_{sa}$ are predicted and modulated by the neural network during the training phase such that they give an estimate of the uncertainty of the model.
Furthermore, the network produces the corresponding state-value estimations ($v_{acc}$ and $v_{sa}$) using two different reward functions $R_{acc,t}$ and $R_{sa,t}$ related to the acceleration and steering angle respectively.

The neural network was trained on the four scenarios contained in the red rectangles of Fig. \ref{fig:hd_map}. For each scenario, we created multiple instances on which the agents act independently of each other. 
Every agent follows the kinematic bicycle model \cite{kong2015kinematic} using values $[-0.2, +0.2]$ for the steering angle and $[-2.0 \frac{m}{s^2}, +2.0\frac{m}{s^2}]$ for the acceleration.
At the beginning of the episode, each agent starts driving with a random speed ($[0.0, 8.0]$) and following its predetermined path and respecting the road speed limits. The road speed limit on this urban area can vary between $4\frac{m}{s}$ to $8.3 \frac{m}{s}$.\\
Finally, since there are no obstacles inside the training scenarios, episodes can finish in one of the following terminal states:
\begin{itemize}
\item \textit{Goal reached}: the final goal position was reached by the agent.
\item \textit{Off-road}: the agent goes outside its predetermined path, predicting erroneously steering angle values.
\item \textit{Time-over}: the time to finish the episode expires; this is mainly due to cautious predictions of the acceleration output, driving the vehicle at lower speeds than the road speed limits.
\end{itemize}

\subsection{Reward}
In order to obtain a policy able to drive the car smoothly both in simulation and in the real world environment, a reward shaping is essential to achieve the desired behavior.
In particular, we define two different reward functions to evaluate the two actions separately: $R_{acc, t}$ and $R_{sa, t }$ related to the acceleration and the steering angle respectively. They can be defined as follows:

\begin{equation}
R_{acc,t} = r_{speed} + r_{acc\_ indecision} + r_{terminal}
\end{equation}
\begin{equation}
R_{sa,t} = r_{localization} + r_{sa\_ indecision} + r_{terminal} 
\end{equation}
$r_{speed}$ is a signal given to $R_{acc, t}$ depending on the value of the ratio ($sr$) between the vehicle current speed and the target speed, and it encourages the agent to achieve but not to exceed the route speed limit, and it can be defined as:
\begin{equation}
r_{speed} = \begin{cases} 
sr \cdot \zeta & \mbox{\textbf{if} } sr < 1.0 ,\\ 
(sr - 1.0) \cdot \zeta & \mbox{\textbf{otherwise},}
\end{cases} 
\end{equation}
where $\zeta$ is a constant set to 0.009.\\
$r_{localization}$ is a penalization given to $R_{sa,t}$ when the position or the heading of the agent differ from those of the road, and it is defined as:

\begin{equation}
r_{localization} = \phi \cdot (h_a - h_p) + \chi \cdot d 
\end{equation}
where $\phi$ and $\chi$ are constants set to 0.05, $h_a$ and $h_p$ are the heading of the agent and that of the road respectively, and finally $d$ is the lateral distance between the position of the agent and the center of the lane.
Both $R_{sa,t}$ and $R_{acc, t}$ have an element in the formula with the purpose of penalizing two consecutive actions that differ by a value greater than a certain threshold, $\delta_{acc}$ and $\delta_{sa}$ for acceleration and steering angle respectively.
In particular, the difference between two consecutive accelerations is calculated as $\Delta_{acc} = |acc(t) - acc(t-1)|$ and $r_{acc\_ indecision}$ is defined as:
\begin{equation}
\label{eq5}
r_{acc\_ indecision} = \psi \cdot \min(0.0, \delta_{acc} - \Delta_{acc}) 
\end{equation}
where $\psi$ is a constant set to 0.1 and $\delta_{acc}$ is set to 0.5 $\frac{m}{s^2}$.\\
Instead, the difference between two consecutive predictions of the steering angle is calculated as $\Delta_{sa} = |sa(t) - sa(t-1)|$, such that $r_{sa \_ indecision}$ is defined as:
\begin{equation}
r_{sa\_ indecision} = \lambda \cdot \min(0.0, \delta_{sa} - \Delta_{sa}) 
\end{equation}
where $\lambda$ is a constant set to 0.01 and $\delta_{sa}$ is set to 0.05.
\\Finally, $R_{acc,t}$ and $R_{sa,t}$ depend on the terminal state achieved by the agent: 
\begin{itemize}
\item \textit{Goal reached}: the agents achieves the goal position, so $r_{terminal}$ is set to $+$1.0 for both rewards.
\item \textit{Off-road}: the agent goes off of its path and it is mainly due to an inaccurate prediction of the steering angle. For this reason we assign a negative signal of $-$1.0 to $R_{sa,t}$ and 0.0 to $R_{acc,t}$.
\item \textit{Time-over}: the available time to finish the episode expires, and this is mainly due to an overly cautious acceleration predictions of the agent; for this reason $r_{terminal}$ assumes the value of $-$1.0 for $R_{acc,t}$ and 0.0 for $R_{sa,t}$. 
\end{itemize}

\subsection{Deep Model}
\label{dm}
One of the main problems related to the use of simulators consists in the difference between simulated and real data, caused by the difficulty of faithfully reproducing real-world situations inside the simulator. 
To overcome this problem we used a synthetic simulator (Fig. \ref{fig:scenario}) in order to simplify the input of the neural network and to reduce the gap between simulated and real world data. Indeed, the information contained in the 4 channels (obstacles, navigable space, path  and stop line) passed as input to the neural network, can be easily reproduced by perception and localization algorithms and by HD Maps embedded on the real self-driving car.

\begin{figure}
    \centering
    \includegraphics[width=0.45\textwidth]{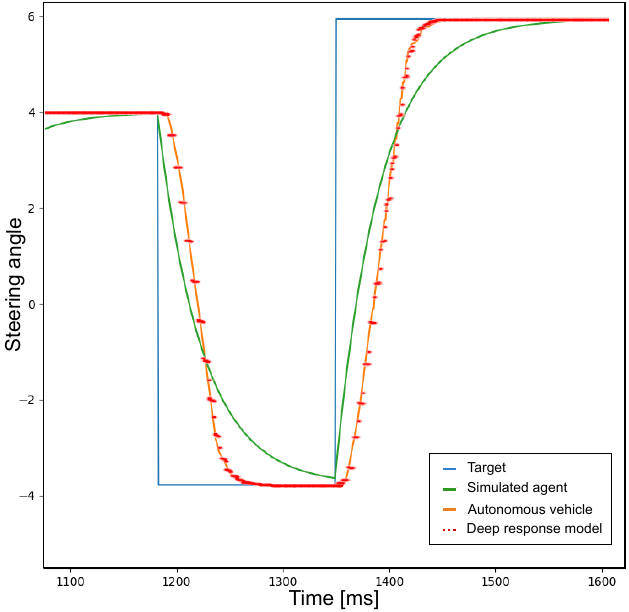}
    \caption{
	Comparison between simulated agents (green curve), real vehicle (orange curve) and \textit{deep\_response} model (red dotted curve) behaviors executing target actions (blue curve).}
    \label{fig:delay}
\end{figure}

Moreover, another relevant problem in the use of a simulator is related to the difference in how simulated agents perform a target action compared to how the autonomous car would behave executing that command. Indeed, a target action computed at time $t$ can ideally be executed with immediate effect at the same precise instant of time in simulation. Differently, this not happens on board of a real vehicle in which such target action will be performed with a certain dynamics, that results in a delay in the execution ($t+\delta$). For this reason, it is necessary to introduce such response time in simulation, in order to train agents to handle such delay on board of the real autonomous car. \\
At this purpose, to achieve a more realistic behavior, we initially trained agents adding a low-pass filter to the target action predicted by the neural network that the agent must perform.
In Fig. \ref{fig:delay}, the blue curve represents the ideal and instantaneous response time that occurs in simulation applying target actions (steering angle in the illustrated example). Then, the green curve identifies the simulated agent response time once a low-pass filter is introduced. Instead, the orange curve shows the behavior of the autonomous vehicle while performing the same steering action. 
However, as we can notice from the figure, the difference in the response time between simulated and real vehicle still remains relevant. \\
Indeed, the acceleration and steering angle set points predicted by the neural network (Fig. \ref{fig:network}) are not feasible commands and do not take into account some factors such as the inertia of the system, the delay of the actuators and other non-idealities.
For this reason, a model consisting in a small neural network composed by 3 fully connected layers (which we will call \textit{deep\_response}) has been developed in order to reproduce as realistically as possible, the dynamics of the real vehicle.
A plot of the \textit{deep\_response} behavior is illustrated by the red dotted curve in Fig. \ref{fig:delay} and it is possible to notice that is very similar to the orange curve which represent that of the real self-driving car. Given the absence of obstacles and traffic vehicles in the training scenarios, the described problem was more evident for the actuation of the steering angle, but the same idea has been applied to the acceleration output. \\
We trained the \textit{deep\_response} model using a dataset collected on board of the autonomous car, in which the inputs correspond to the commands given to the vehicle by a human driver (accelerator pressure and steering wheel movements) and the output correspond to the throttle, brake and curvature of the vehicle that can be measured with GPS, odometry or other techniques.
In this way, we embedded such model in the simulator in order to obtain a more scalable system aims at reproducing the self-driving car behavior.
Therefore, the \textit{deep\_response} module was essential for the correction of the steering angle but even, if in a less evident way, it is also necessary for the acceleration and this will be clearly visible with the introduction of obstacles.

\section{EXPERIMENTS}

To validate the proposed system, some experiments were conducted both in simulation and in the real world. The input of the neural network used in the autonomous vehicle can be obtained with perception algorithms and through HD Maps; in this way, it is therefore possible to reconstruct the same input used in simulation.\\
Initially, we tested two different policies on real data to verify the impact of the \textit{deep\_response} model on the system. Subsequently, we verified that the vehicle followed the path correctly and that it respected the speed limits retrieved by the HD Maps. Finally, we demonstrate that a pre-training of the neural network with Imitation Learning drastically reduces the overall training time.

\subsection{Test on Real Data}

\begin{figure*}[bp]
    \centering
    \subfloat[Path with different speed limits]{\includegraphics[width=0.5\textwidth]{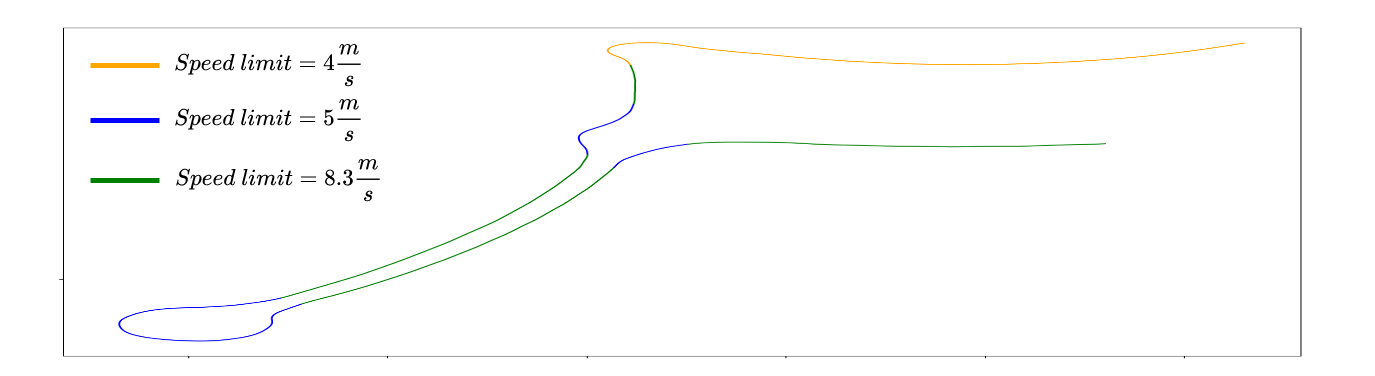}\label{speed_limit_a}}\\
    \subfloat[Acceleration]{\includegraphics[width=0.50\textwidth]{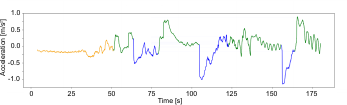}\label{speed_limit_b}}\hfill
    \subfloat[Speed]{\includegraphics[width=0.50\textwidth]{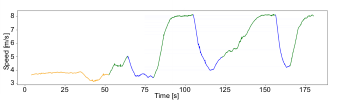}\label{speed_limit_c}}
    \caption{Fig. \ref{speed_limit_a} represents the path performed with the self-driving car in which colors identify the different speed limits faced in this area. Fig. \ref{speed_limit_b} and Fig. \ref{speed_limit_c} are respectively the trend of acceleration and speed during the real test. }
    \label{fig:speed_limit}
\end{figure*}

\begin{figure}
    \centering
    \includegraphics[width=1.0\linewidth]{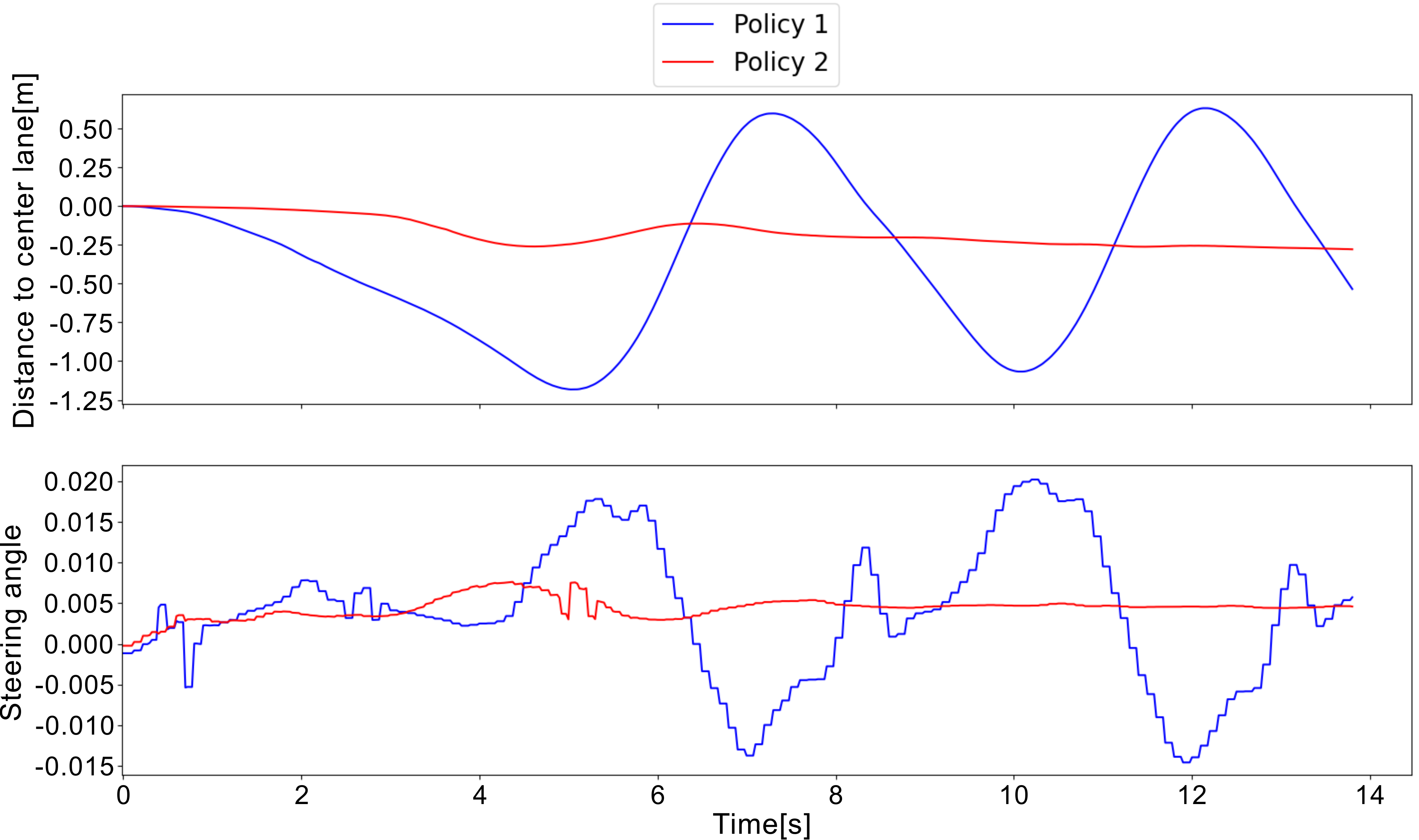}
    \caption{The distance to the center lane and the steering angle output predicted by \textit{Policy 1} (blue curves) and \textit{Policy 2} (red curves) on a short time window of the real world test on board the self-driving car.}
    \label{fig:policies}
\end{figure}

We analyze the behavior of the vehicle over the entire mapped area with two different policies:
\begin{itemize}
\item \textit{Policy 1}: trained without using the \textit{deep\_response} model, but simulating the response of the real vehicle to a target action with a low-pass filter.
\item \textit{Policy 2}: trained by introducing the \textit{deep\_response} model to ensure a more realistic dynamics.
\end{itemize}
The tests performed in simulation led to excellent results with both policies. Indeed, the agent was able to reach the goal in 100$\%$ of cases with a smooth and safe behavior, both in training scenarios and in those parts of the mapped area not contained in the training environments.\\
Different results were obtained by testing the policies on real world scenarios.
\textit{Policy 1} is not able to handle the vehicle dynamics, which will execute the predicted actions differently compared to agent in simulation; in this way, \textit{Policy 1} will observe unexpected states as a consequence of its predictions, causing noisy and uncomfortable behavior on board of the self-driving car.\\
This behavior also affects the reliability of the system, indeed sometimes it was necessary an human assistance to avoid the autonomous car going out of road.\\
On the contrary, \textit{Policy 2} has never required human assistance during all the real world tests performed on self-driving car since it is aware of the vehicle dynamics and so how the system will evolve to the predicted actions. The only cases in which a human interventions has been required is to avoid other road users; however, we did not consider these cases as failures since both \textit{Policy 1} and \textit{Policy 2} are trained in obstacle-free scenarios. 

To better understand the difference between  \textit{Policy 1} and \textit{Policy 2} we plot (Fig. \ref{fig:policies}) the steering angles predicted by the neural network and the distance to the center lane in a short time window of the real world tests. We can notice how the behavior of the two policies are completely different noticing that the \textit{Policy 1} (blue curve) is noisy and unsafe compared to the \textit{Policy 2} behavior (red curve), proving that the \textit{deep\_response} module is essential for the deployment of the policy on board of the real self-driving car.

An example of the real world test performed using the \textit{Policy 2} is illustrated in video\footnote[1]{\url{https://drive.google.com/file/d/1ZCI2qqNPY2CsE4U1xKfEqGzNintDxf9Q/view?usp}}.

\subsection{Speed Limits}

\begin{figure*} [h!]
    \centering
    \subfloat[Positive episodes]{\includegraphics[width=0.49\textwidth]{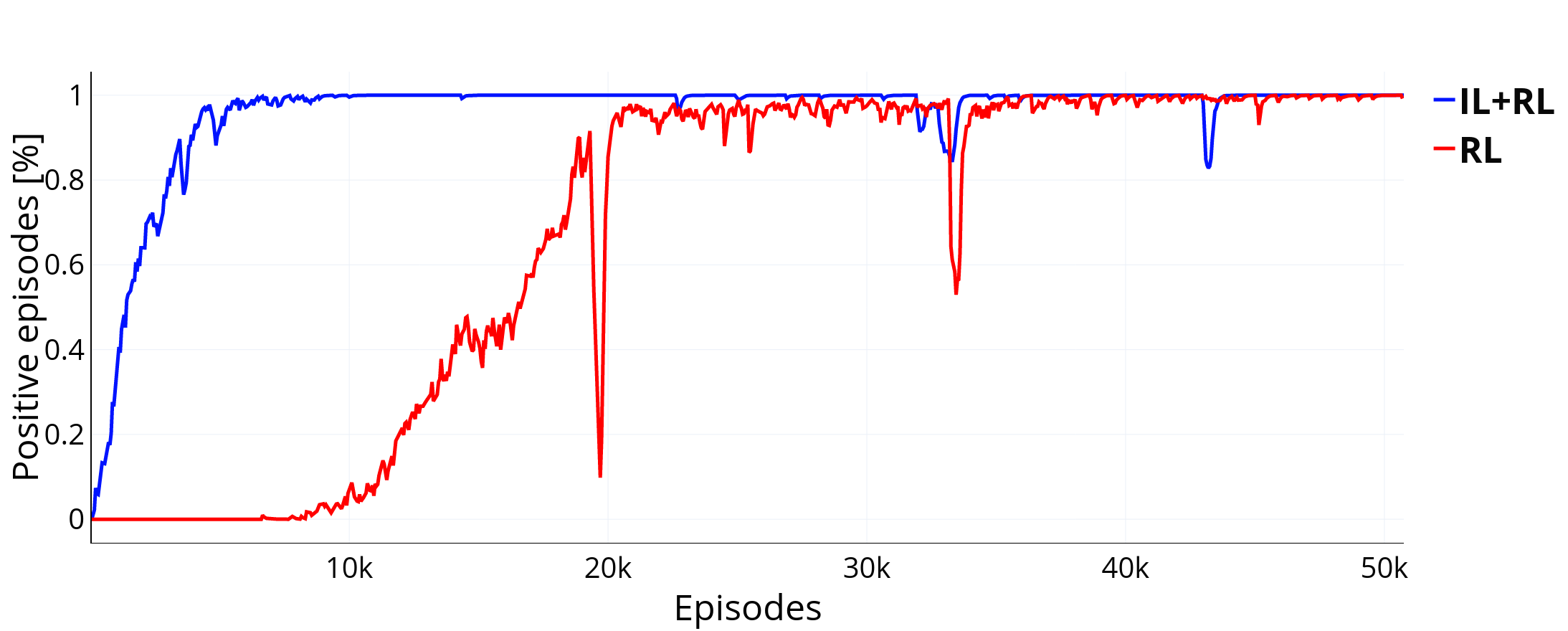}\label{imit_a}}
    \subfloat[Average reward]{\includegraphics[width=0.49\textwidth]{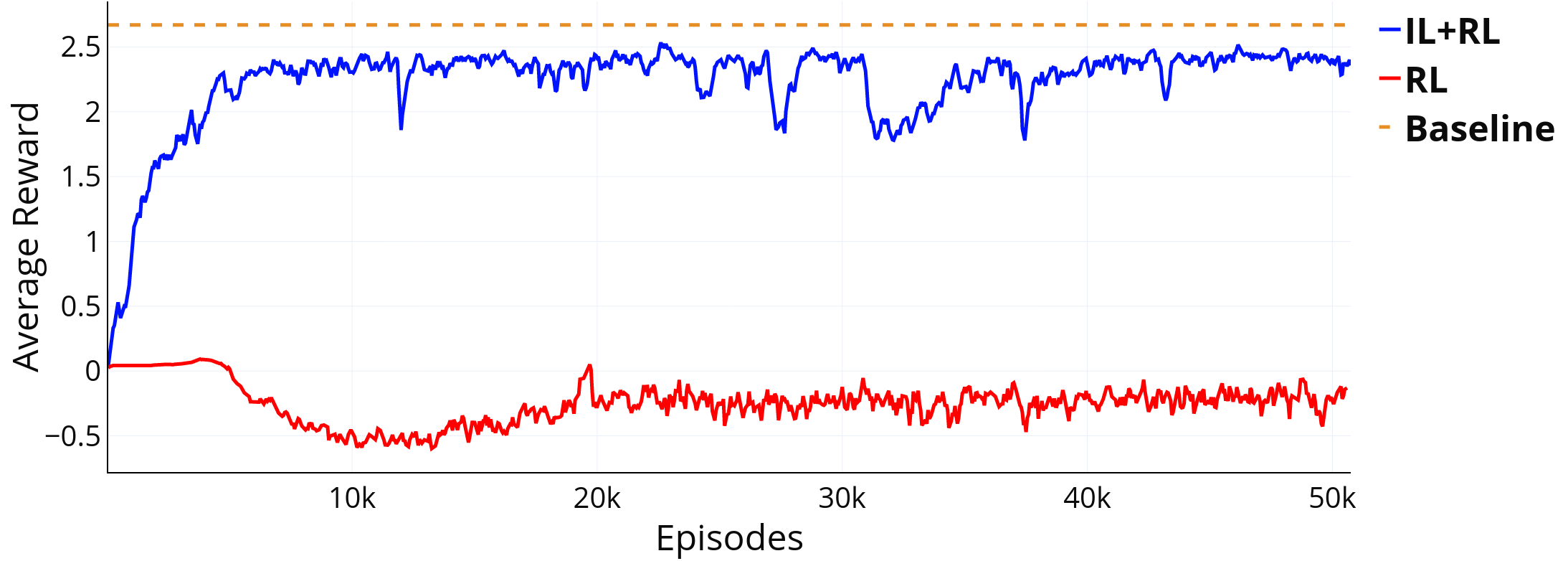}\label{imit_b}}
    \caption{Comparison between \textit{Pure RL} training (red curve) and using a pre-training through Imitation Learning, \textit{IL + RL} (blue curve). In Fig. \ref{imit_b} the orange dashed line represents the average rewards obtained in 50000 episodes using simulated agents that follow deterministic rules as those used for collecting the dataset for IL pre-training.}
    \label{fig:imit}
\end{figure*}
In the second part of the work we verify that the vehicle was able to follow the path and maintain the speed limits.
In Fig. \ref{fig:speed_limit} we illustrate the behavior of the vehicle trained with \textit{Policy 2} in a real urban area. In particular, the Fig. \ref{speed_limit_a} represents a portion of the map (Fig. \ref{fig:hd_map}) with the speed limits of 4$\frac{m}{s}$, 5$\frac{m}{s}$ and 8.3$\frac{m}{s}$, corresponding to orange, blue and green curve colors respectively.
In order to show a more detailed example of the longitudinal behavior of the real self-driving car, we plot the accelerations predicted by the network (Fig. \ref{speed_limit_b}) and the vehicle speeds (Fig. \ref{speed_limit_c}) obtained driving along the route of Fig. \ref{speed_limit_a}.
The results shown in the plots proving that the system is able to maintain the speed limits retrieved by the HD Maps.
Looking at the trend of the acceleration curve in Fig. \ref{speed_limit_b}, it is also important to notice that even if the acceleration output seems noisy, the difference between two consecutive predicted values never exceeds the threshold $\delta_{acc}$ used in the reward function in Equation \ref{eq5}, resulting in a smooth and comfortable longitudinal behavior perceived on-board vehicle.

\subsection{Imitation Learning Pre-training}

In order to overcome a known limit of RL that is the need for millions of episodes to reach the optimal solution, we performed a pre-training through Imitation Learning.
Moreover, even if the trend in using IL is to train large models, we used the same small neural network (Fig. \ref{fig:network}) (about a million parameters), since the idea is to continue training the system using the RL framework in order to ensure more robustness and generalization capabilities. In this way, we do not increase the hardware resources usage and it is essential considering possible future multi-agent trainings \cite{dorri2018multi}. 

The dataset used during the IL training phase is generated by simulated agents that move following rule-based approaches.
In particular, for the curvature we use the tracking algorithm called Pure Pursuit \cite{coulter1992implementation}, where the agent's purpose is to move following specific waypoints. Instead, we use the Intelligent Driver Model (IDM) \cite{treiber2000congested} to control the longitudinal acceleration of the agent. \\
To create the dataset, the rule-based agents move on the four training scenarios (Fig. \ref{fig:hd_map}) and every 100 ms they save the scalar parameters and the four visual input. Instead, the output is given by the Pure Pursuit algorithm and the Intelligent Driver Model.\\
The two lateral and longitudinal controls, which correspond to the output, only represent the tuple ($\mu_{acc}$, $\mu_{sa}$). Consequently, during the IL training phase we do not estimate the values of the standard deviations ($\sigma_{acc}$, $\sigma_{sa}$) and neither the value functions ($v_{acc}$, $v_{sa}$). These features, together with the \textit{deep\_response} module,  are learned during the \textit{IL + RL} training phase.\\
In Fig. \ref{fig:imit} we show the comparison of the results obtained by training the same neural network starting with the pre-training phase (blue curve, \textit{IL + RL}) and using only Reinforcement Learning (red curve, \textit{Pure RL}) on all the four scenarios (Fig. \ref{fig:hd_map}).
Both approaches achieve good success rates (Fig. \ref{imit_a}), even if \textit{IL + RL} training requires less episodes than the \textit{Pure RL} and the trend is also more stable. Furthermore, the reward curves illustrated in Fig. \ref{imit_b} prove that the policy obtained with the \textit{Pure RL} approach (red curve) did not even achieve an acceptable solution requiring more training time, while the \textit{IL + RL} policy reaches an optimal solution in few episodes (blue curve in Fig. \ref{imit_b}). 
In this case an optimal solution is represented by the dashed orange curve. This baseline represents the average rewards obtained using simulated agents that perform 50000 episodes in the 4 scenarios of Fig. \ref{fig:hd_map}. The simulated agents follow deterministic rules as those used for collecting the dataset for IL pre-training, which therefore use Pure Pursuit for curvature and IDM for longitudinal acceleration. 
This gap between the two approaches may be more evident training the system performing more complex maneuvers in which agents interactions could be required.

\section{CONCLUSIONS}

In this paper we developed a planner based on Reinforcement Learning that allows the vehicle to drive safely and comfortably in an urban environment. The agents in simulation were trained using a neural network that predict acceleration and steering angle each 100 ms. 
However, at this stage of work, we used obstacle-free training environments in which agents were only trained to follow their paths and to observe the road speed limits.\\
We also presented the \textit{deep\_response} system that allows to 
reproduce the real vehicle behavior in simulation.
In particular, the simulated agents learn to perform target actions evolving their states with a dynamics as similar as possible to that of the real vehicle. This allowed us to achieve a more comfortable driving style testing the system on real scenarios. \\
Currently, no studies have been conducted to verify the generalization capability of the \textit{deep\_response} model, because a fleet of different autonomous vehicles would be required to test the system.

With the current system we performed real tests in the mapped area (Fig. \ref{fig:hd_map}), that correspond to a low-traffic neighborhood of Parma. 
We observed that the system reached good performances in the entire area and was able to drive smoothly both in those parts used as training scenarios and in unknown ones, proving that did not overfit on the only training
environments, but it have a good generalization capability. \\
Finally, we have also shown that with a pre-training using Imitation Learning, we can drastically reduce training times compared to the pure Reinforcement Learning.

Future developments of this work are aimed at creating a multi-agent system in which the vehicle is able to perform other types of maneuvers, such as Adaptive Cruise Control, obstacle avoidance or intersection handling.\\
Furthermore, an interesting future development would be to solve the current limit on the size of visual inputs and consequently of their content. Currently the visual inputs (Fig. \ref{fig:scenario}) contain a surrounding of 50$\times$50 meters of the agent and this allows us to represent only an area of 40 m forward and 25 m on the sides.
However, choosing a larger visual input definitely requires a more complex feature extractor than the current one. A possible solution could be the Variational Autoencoder (VAE) \cite{kingma2013auto}, but also in this case we will have to be careful to balance the system to avoid too long training times and computationally too expensive processes.






\bibliography{root}
\bibliographystyle{ieeetr}

\end{document}